\documentclass[11pt,a4paper]{article}
\usepackage[hyperref]{emnlp2020}
\usepackage{times}
\usepackage{latexsym}
\usepackage{url}
\usepackage{booktabs}       
\usepackage{amsfonts}       
\usepackage{nicefrac}       
\usepackage{microtype}      
\usepackage{amssymb}
\usepackage{amsmath, calc}
\usepackage{bm}
\usepackage{multirow}
\usepackage{multicol}
\usepackage{graphicx}
\usepackage{courier}
\usepackage{wrapfig}
\usepackage{adjustbox}
\usepackage{tabularx}
\usepackage{diagbox}
\usepackage{makecell}
\usepackage{caption}
\usepackage{enumitem}
\usepackage{titlesec}
\usepackage{soul}
\usepackage{enumitem}
\usepackage[hang]{footmisc}

\usepackage{letltxmacro}
\usepackage{pbox}
\usepackage{etoolbox}
\usepackage{blindtext}
\usepackage{subcaption}

\usepackage{textcomp}
\usepackage{filecontents}
\newcommand{\pnt}[1]{{\scriptstyle#1}}

\usepackage{microtype}
\aclfinalcopy 
\def\aclpaperid{2889} 

\title{Joint Turn and Dialogue level User Satisfaction Estimation on Multi-Domain Conversations}

\author{
Praveen Kumar Bodigutla \\
  \texttt{pbodigutla@linkedin.com\thanks{Currently at LinkedIn, but did this work at Amazon.}}
  \\\And
Aditya Tiwari\\
\texttt{aditiwar@amazon.com} 
  \\\AND
Josep Valls Vargas\\
 \texttt{jvalls@amazon.com} 
  \\\And
Lazaros Polymenakos\\
 \texttt{polyml@amazon.com} 
  \\\And
Spyros Matsoukas\\ 
 \texttt{matsouka@amazon.com} 
 }

\date{}

\begin{document}
\maketitle

\begin{abstract}
Dialogue level quality estimation is vital for optimizing data driven dialogue management. Current automated methods to estimate turn and dialogue level user satisfaction employ hand-crafted features and rely on complex annotation schemes, which reduce the generalizability of the trained models. We propose a novel user satisfaction estimation approach which minimizes an adaptive multi-task loss function in order to jointly predict turn-level Response Quality labels provided by experts and explicit dialogue-level ratings provided by end users. The proposed BiLSTM based deep neural net model automatically weighs each turn's contribution towards the estimated dialogue-level rating, implicitly encodes temporal dependencies, and removes the need to hand-craft features. 

On dialogues sampled from $28$ Alexa domains, two dialogue systems and three user groups, the joint dialogue-level satisfaction estimation model achieved up to an absolute 27\% (0.43 $\rightarrow$ 0.70) and 7\% (0.63 $\rightarrow$ 0.70) improvement in linear correlation performance over baseline deep neural net and benchmark Gradient boosting regression models, respectively. 
\end{abstract}

\section{Introduction}

\label{sec:intro}
Automatic turn and dialogue level quality evaluation of end user interactions with Spoken Dialogue Systems (SDS) is vital for identifying problematic conversations and for optimizing dialogue policy using a data driven approach, such as reinforcement learning. One of the main requirements to designing data-driven policies is to automatically and accurately measure the success of an interaction. Automated dialogue quality estimation approaches, such as Interaction Quality (IQ)   \citep{SCHMITT12.333} and recently Response Quality (RQ) \citep{Bodigutla_SIGDIAL2019} were proposed to capture satisfaction at turn level from an end user perspective. Automated models to estimate IQ \citep{10.1007/978-1-4614-8280-2_27, Schmitt2011ModelingAP, 6854195} used a variety of features derived from the dialogue-turn, dialogue history, and output from three Spoken Language Understanding (SLU) components, namely: Automatic Speech Recognition (ASR), Natural Language Understanding (NLU), and the dialogue manager. RQ prediction models \citep{Bodigutla_SIGDIAL2019} further extended the feature sets with features derived from the dialogue-context, aggregate popularity and diversity of topics discussed within a dialogue-session. 

Using automatically computed diverse feature sets and expert ratings to annotate turns overcame limitations suffered by earlier approaches to measure dialogue quality at turn-level, such as using sparse sentiment signal \citep{Shi2018SentimentAE}, intrusive solicitation of user feedback after each turn, and using manual feature extraction process to estimate turn-level ratings \citep{engelbrecht-etal-2009-modeling, Higashinaka2010IssuesIP}.

For predicting user satisfaction at dialogue-level, IQ estimation approach was shown to generalize to dialogues from different domains \citep{article_Ultes_speechcom}. Using annotated user satisfaction ratings to estimate dialogue-level quality, overcame the limitation with using task success  \citep{schatzmann-etal-2007-agenda} as dialogue evaluation criteria. Task success metric does not capture frustration caused in intermediate turns and assumes the end user goal is known in advance. However, IQ annotation approach to rate each turn incrementally, lowered Inter Annotator Agreement (IAA) for multi-domain dialogues \citep{Bodigutla_SNEURIPS2019}. Multi-domain dialogues are conversations that span multiple domains (Table \ref{tab:multi-domain-example}) in a single dialogue-session. On the contrary, RQ ratings were provided for each turn independently and were shown to be highly consistent, generalizable to multiple-domain conversations and were highly correlated with turn-level explicit user satisfaction ratings \citep{Bodigutla_SNEURIPS2019}.  Furthermore, using predicted turn-level RQ ratings as features, end-user explicit dialogue-level ratings for complex multi-domain conversations were accurately predicted across dialogues from both new and seasoned user groups \citep{Bodigutla_SNEURIPS2019}. Earlier widely used approach, such as PARADISE \citep{Walker:2000:TDG:973935.973945}, where the model is trained using noisy end dialogue ratings provided by users, did not generalize to diverse user population \citep{DBLP:journals/corr/abs-1905-04071}.

Despite generalizing to different user groups and domains, both turn and dialogue level quality estimation models trained using annotated RQ ratings \citep{Bodigutla_SIGDIAL2019, Bodigutla_SNEURIPS2019} used automated, yet hand-crafted features. Modern day SDS support interoperability between different dialogue systems, such as ``pipeline based modular'' and ``end-to-end neural'' dialogue systems. Hand-crafted features designed based on one system are not guaranteed to generalize to dialogues on a new system. 

RQ based dialogue-level satisfaction estimation models \citep{Bodigutla_SNEURIPS2019} did not factor in noise in explicit user ratings and used average estimated turn-level RQ ratings as a feature to train the model. Each turn's success or failure was assumed to have an equal contribution to the overall dialogue rating. However, a user might be dissatisfied even if most of the turns in the dialogue were successful (example in Appendix Table \ref{tab:unequal-weighting}). 

The LSTM \citep*{Hochreiter:1997:LSM:1246443.1246450} based IQ estimation approaches \citep{Pragst2017, Rach2017InteractionQE} were shown to encode temporal dependencies between turns implicitly. Most recently, BiLSTMs (Bi-directional LSTMs) with self-attention mechanism \citep{ultes-2019-improving}, which used only turn-level features achieved best performing IQ estimation performance. 

In order to address the aforementioned limitations with using hand-crafted features, we propose a LSTM \citep*{Hochreiter:1997:LSM:1246443.1246450} based turn-level RQ estimation model, which implicitly encodes temporal dependencies and removes hand-crafting of turn and temporal features. Along with turn-level features that are not dialogue-system or user group specific, we use features derived from pre-trained Universal Sentence Encoder (USE) embeddings \citep{DBLP:journals/corr/abs-1803-11175} of an utterance and system response texts to train the model. Pre-trained sentence representations provided by USE Transformer model achieved excellent results on semantic relatedness and textual similarity tasks \citep{DBLP:journals/corr/abs-1806-06259}.  

Using an adaptive multi-task loss weighting technique \citep{DBLP:journals/corr/KendallGC17} and attention \citep{DBLP:journals/corr/VaswaniSPUJGKP17} over predicted turn-level ratings, we further extend the turn-level model to design a novel BiLSTM \citep{Graves13hybridspeech} based joint turn and dialogue-level quality estimation model. To test the generalization performance of the proposed approaches, we estimate turn and dialogue-level ratings on multi-turn \footnote{In single-turn conversations the entire context is expected to be present in the same turn. In multi-turn case context from previous turns is carried to address user’s current request.} multi-domain conversations sampled from three user groups, spanning $28$ domains  (e.g., \textit{Music, Weather, Movie \& Restaurant Booking}) across two different dialogue systems.

To the best of our knowledge, this is the first attempt to leverage noise adaptive multi-task deep learning approach to jointly estimate annotated turn-level RQ and user provided dialogue level ratings for multi-domain conversations from multiple user groups and dialogue systems.

The outline of the paper is as follows: Section 2 discusses the choice of RQ annotation. Section 3 \& 4 presents the novel approaches to estimate turn and dialogue level quality ratings. Section 5 summarizes the turn and dialogue level data and presents our experimental setup. Section 6 provides an empirical study of the models' performance. Section 7 concludes.

\begin{table}[t]
\centering
\small
\resizebox{0.31\textheight}{!}{
 \begin{adjustbox}{max width=1.0\textwidth}
  \bgroup
  \def\arraystretch{1.4}
\begin{tabular}{p{5.8cm}p{4.4cm}}
\hline
\textbf{User utterance \{Domain\}} & \textbf{System Response} \\
\hline
\hline
Movies playing near me \{MovieBooking\} &  Avengers is playing at AMC \\
What are the showtimes? \{MovieBooking\} & 4 p.m and 7 p.m \\
Two tickets for 7 p.m show \{MovieBooking\} & Booking 2 tickets for 7 p.m ...\\
Is it raining today? \{Weather\} & Yes there is a 60\% chance of rain\\
Book cab to the theater at 6 p.m \{CabBooking\} & Ride for 2 costs \$20, should I book? \\
Yes \{CabBooking\} & Sure, booked a cab for 2 to AMC \\
Cancel 7 p.m appointment \{Calendar\} & Canceled your 7 p.m park visit\\
\hline
\end{tabular}
\egroup
\end{adjustbox}
}
\caption{Example multi-domain dialogue a user has with the conversation agent to plan his/her evening. The conversation spans $4$ domains in a single dialogue session.}
\vspace{-0.5cm}
  \label{tab:multi-domain-example}
\end{table}

\section{Response Quality for Turn and Dialogue level Quality Estimation}
\label{sec:rq}
Interaction Quality (IQ) \citep{SCHMITT12.333}  and and Task Success (TS) \citep{schatzmann-etal-2007-agenda} measures require an annotator to accurately determine the task that the user is aiming to accomplish through a dialogue, which is non-trivial for multi-domain conversations  \citep{Bodigutla_SNEURIPS2019}. Both IQ and RQ \citep{Bodigutla_SIGDIAL2019} require annotators to rate each turn on a discrete five point scale (RQ rating scale in Appendix Table \ref{tab:rq-scale}). Unlike IQ, RQ annotators need not keep track of dialogue progression so far to rate an individual turn.  Due to the simplicity of annotation scheme, multi-domain generalizability and applicability to dialogue level satisfaction estimation \citep{Bodigutla_SNEURIPS2019}, we chose turn-level RQ annotation scheme. Similar to \citep{Bodigutla_SNEURIPS2019}, the dialogue-level ratings are directly obtained from end users who interacted with different dialogue systems. Unlike TS metric,  which does not capture user's dissatisfaction in intermediate turns, dialogue-level satisfaction ratings holistically capture the overall satisfaction of an end user's interaction with SDS.

\section{Turn-level Dialogue Quality Estimation}
\label{sec:turn-level}
In this section we discuss previous turn-level satisfaction estimation models trained using RQ ratings, their limitations and our approach to overcome them.

\begin{figure}[!h]
  \begin{center}
    \includegraphics[width=0.40\textwidth]{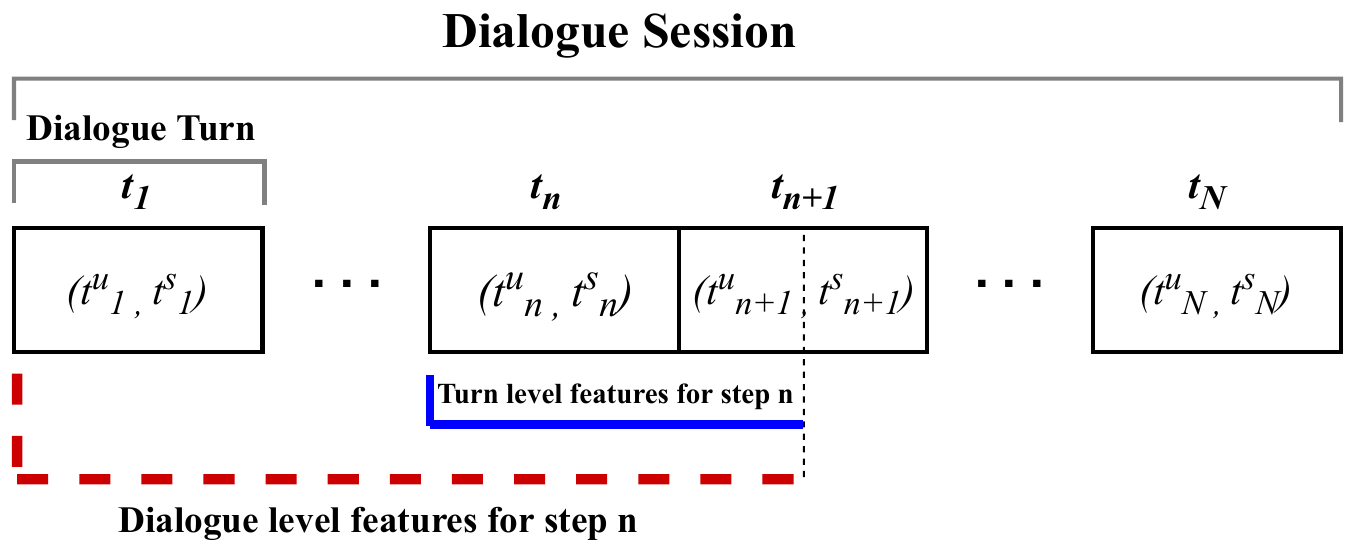}
  \end{center}
  \vspace{-0.2cm}
  \caption{Dialogue and turn definitions for estimating user satisfaction rating on turn $t_n$ \cite{Bodigutla_SIGDIAL2019}.  The solid blue and dotted red lines indicate the context used for generating turn and dialogue level features respectively.}
  \label{fig:turn-dialog-parameters}
\vspace{-0.2cm}
\end{figure}

Similar to \citet{Bodigutla_SIGDIAL2019}, we define a dialogue turn at time $n$ as $t_n=(t_n^u, t_n^s)$, where $t_n^u$ and $t_n^s$ represent the user request and system response on turn $n$ respectively (Figure \ref{fig:turn-dialog-parameters}). A dialogue session of $N$ turns is defined as ($t_1$:$t_N$). In experiments conducted by \citet{Bodigutla_SIGDIAL2019}, Gradient Boosting Regression \citep{friedman2001greedy} model gave the best turn-level RQ prediction performance. Features used to train the model were derived from current turn ($t_{n}$), dialogue history ($t_{1:n-1}$) and next turn's user request ($t_{n+1}^{u}$). In addition to deriving domain-independent features from three SLU components, namely Automatic Speech Recognition (ASR), Natural Language Understanding (NLU), and the dialogue manager, five new feature sets were introduced by the authors to improve the performance of the turn-level satisfaction estimation model.
\vspace{0.1cm}

Features used in the model were automatically computed, yet they were carefully hand-engineered (See Appendix Table \ref{tab:features-hand-craft}). Features were hand-crafted to identify and rank factors contributing to the predicted satisfaction rating, but these features do not generalize easily to different dialogue systems. Introduced originally by authors of RQ, ``un-actionable request'' feature was computed by identifying the presence of particular key words (e.g., ``sorry'', ``i don't know'') in the system's response. This rule-based feature does not generalize to a system that uses different set of phrases to indicate its inability to satisfy user's request. Even temporal dialogue level features computed over turns ($t_1$:$t_n$) were also hand-crafted and computed by taking simple aggregate statistics (e.g., mean) over turn level features. 

\subsection{LSTM-based Response Quality Estimation Models}
\label{subsec:turn-level-lstm}
\begin{figure}[!h]
  \begin{center}
    \includegraphics[width=0.35\textwidth]{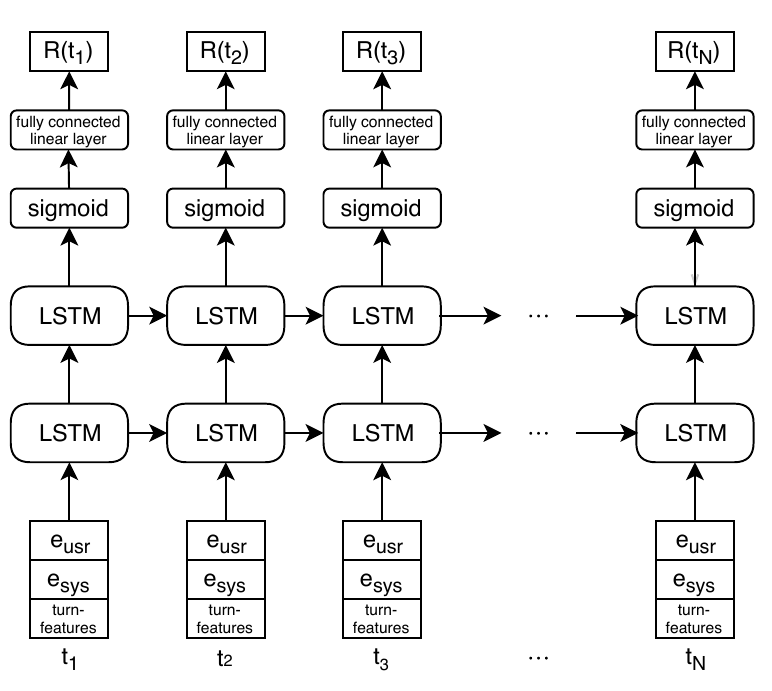}
  \end{center}
\vspace{-0.1cm}
  \caption{Uni-directional LSTM model to predict RQ ratings at each time-step to estimate dialogue quality at turn-level. $e_{usr}$, $e_{sys}$ and $turn{-}features$ are pre-trained Universal Sentence Encoder embeddings for user request,  system response and rest of the features in Table \ref{tab:turn-level-features} respectively.}
  \label{fig:turn-model}
\end{figure}

In order to overcome the limitation of hand-crafting temporal features, we propose using a Long Short Term Memory (LSTM) \citep*{Hochreiter:1997:LSM:1246443.1246450} based model to estimate turn-level satisfaction ratings sequentially on a continuous $[1{-}5]$ scale. \citet{Rach2017InteractionQE} showed that by using only turn-level features, pre-computed temporal features were no longer required for estimating IQ using a LSTM network. To keep the turn-level dialogue quality estimation system causal \citep*{litan_2013}, where the output at the current time step only depends on current and previous steps, we do not introduce bi-directionality \citep{Graves13hybridspeech} into the network architecture (See Figure \ref{fig:turn-model}). Unlike dialogue-level rating, which is computed at the end of a dialogue-session, only past dialogue-context is available to compute a turn's quality rating. Causality enables using turn-level model to optimize dialogue policies online.

Models encoding sentences into embedding vectors have been successfully used in transfer learning and performing several downstream Natural Language Processing tasks (e.g., Classification and semantic textual similarity detection). Pre-trained sentence representations provided by Universal Sentence Encoder (USE) \citep{DBLP:journals/corr/abs-1803-11175} model achieved excellent results on semantic relatedness and textual similarity tasks \citep{DBLP:journals/corr/abs-1806-06259}. 

To address the limitation with using features derived from hand-crafted rules, we use feature sets which are derived from USE pre-trained (512 dimensional) embeddings from its transformer variant.  We introduce a set of five features derived from USE embeddings of user request and system response texts (See Table {\ref{tab:turn-level-features}). These features are then concatenated with turn-level features obtained from the SLU (e.g., ASR confidence score), dialogue manager (e.g., system response) output and predicted intent and domain popularity statistics. Concatenated features are passed as input to each time-step of the uni-directional turn-level satisfaction estimation deep LSTM network (Figure \ref{fig:turn-model}), that minimizes mean square error loss between actual and predicted turn-level RQ ratings. 



\begin{table}[t]
\centering
\resizebox{0.31\textheight}{!}{
 \begin{adjustbox}{max width=1.0\textwidth}
  \bgroup
  \def\arraystretch{1.4}
\begin{tabular}{ll}
\hline
Feature name & Methodology used to compute the feature\\
\hline
\hline
ASR Conf. score  &  Available in the output of the ASR system\\
NLU Conf. score   &  Available in the output of the NLU system\\
Barge-in & Output from ASR \\
\textbf{USE embedding of user request }& USE embeddings of $t_{n}^u$  \\
\textbf{USE embedding of system response} & USE embeddings of $t_{n}^s$ \\
NLU intent similarity & Sim. between NLU predicted intents for $t_{n}^u$ and  $t_{n+1}^u$\\
\textbf{Semantic paraphrase of user req.} &  Cosine sim. between USE embeddings of $t_{n}^u$ \& $t_{n+1}^u$ \\ 
Syntactic paraphrase of user req. &  Jaccard sim. between words in $t_{n}^u$ \& $t_{n+1}^u$\\
\textbf{Semantic req. \& resp. coherence} & Cosine sim. between USE embeddings of $t_{n}^u$ \& $t_{n}^s$ \\
Syntactic req. \& resp. coherence &  Jaccard sim. between words in $t_{n}^u$ \& $t_{n+1}^s$\\
\textbf{Semantic resp. repetition }& Cosine sim. between USE embeddings of $t_{n}^s$ \& $t_{n-1}^s$ \\
Syntactic resp. repetition & Jaccard sim. between words in $t_{n}^s$ \& $t_{n-1}^s$ \\
Length of User utterance & Number of words in $t_{n}^u$\\
Length of resp. & Number of words in $t_{n}^s$\\
Duration between utterances & Seconds elapsed between  $t_{n}^u$ \& $t_{n+1}^u$ \\
Domain popularity &  Avg. \# of reqs. per user for predicted NLU domain $t_{n}^u$ \\
Intent popularity & Avg. \# of reqs. per user for predicted NLU intent $t_{n}^u$ \\
\hline
\end{tabular}
\egroup
\end{adjustbox}
}
\caption{Turn level features for turn-${t_{n}}$ and the methodology used to compute them. In bold are features derived from USE embeddings that we introduced. Rest of the turn-level features are similar to \citet{Bodigutla_SIGDIAL2019} (Appendix Table \ref{tab:features-hand-craft}). Note {\scriptsize $\sim$}65\% relative drop in number of features ($48$$ \rightarrow$ $17$). resp., conf., avg., sim., \#, \& req. indicate response, confidence, average, similarity, count and request respectively. }
\vspace{-0.6cm}
  \label{tab:turn-level-features}
\end{table}

\section{Dialogue-level Quality Estimation}
\label{sec:dialogue-level}
In this section we discuss the novel joint turn and dialogue quality estimation approach.

\subsection{Joint Estimation of Turn and Dialogue Level Ratings}
\label{sec:dialogue-level-lstm}
Turn-level satisfaction estimation helps identify a particular turn's success from an end user's perspective. In addition to predicting whether individual turn was successful, we need a dialogue level user satisfaction metric for learning dialogue policies that maximize end user satisfaction on the overall dialogue. Dialogue-level metric also helps in identifying problematic dialogues which caused dissatisfaction to the end user. 

We propose a novel approach (Figure \ref{fig:dialogue-model}) to jointly predict turn and dialogue level satisfaction ratings for a given dialogue. Unlike turn-level satisfaction estimation, we are not constrained to use only historical context of a dialogue to predict the dialogue-level ratings as entire context of the dialogue is available while predicting a dialogue level rating. Hence instead of LSTMs we use deep BiLSTM \citep{Graves13hybridspeech} network for the dialogue-level satisfaction estimation task. \citet{ultes-2019-improving} showed that BiLSTMs with self-attention \citep{Zheng:2018:OOA:3219819.3219839} model gave the best performance on the IQ prediction task and the model implicitly encoded temporal dependencies. Feature inputs to the joint model are same as the ones we use for turn-level quality estimation in Section \ref{subsec:turn-level-lstm}.

Individual turn's predicted RQ rating does not provide enough information to estimate whether an entire dialogue is satisfactory. \citet{Bodigutla_SNEURIPS2019} used average turn-level predicted RQ ratings as feature to estimate dialogue-level quality. We hypothesize that users do not equally weigh each each turn's success (or failure) while determining end dialogue rating (Example conversation in Appendix Table \ref{tab:unequal-weighting}). We apply attention \citep{DBLP:journals/corr/VaswaniSPUJGKP17}  over turn-level ratings and concatenate the aggregate weighted turn-level rating with the entire dialogue's representation (hidden state $h_{t_{N}}$ in Figure \ref{fig:dialogue-model}) before passing it through the sigmoid activation layer for dialogue rating prediction.

In the next section we describe the multi-task loss function we minimized for jointly estimating turn and dialogue-level quality ratings.

\begin{figure}[!h]
  \begin{flushleft}
    \includegraphics[width=0.45\textwidth]{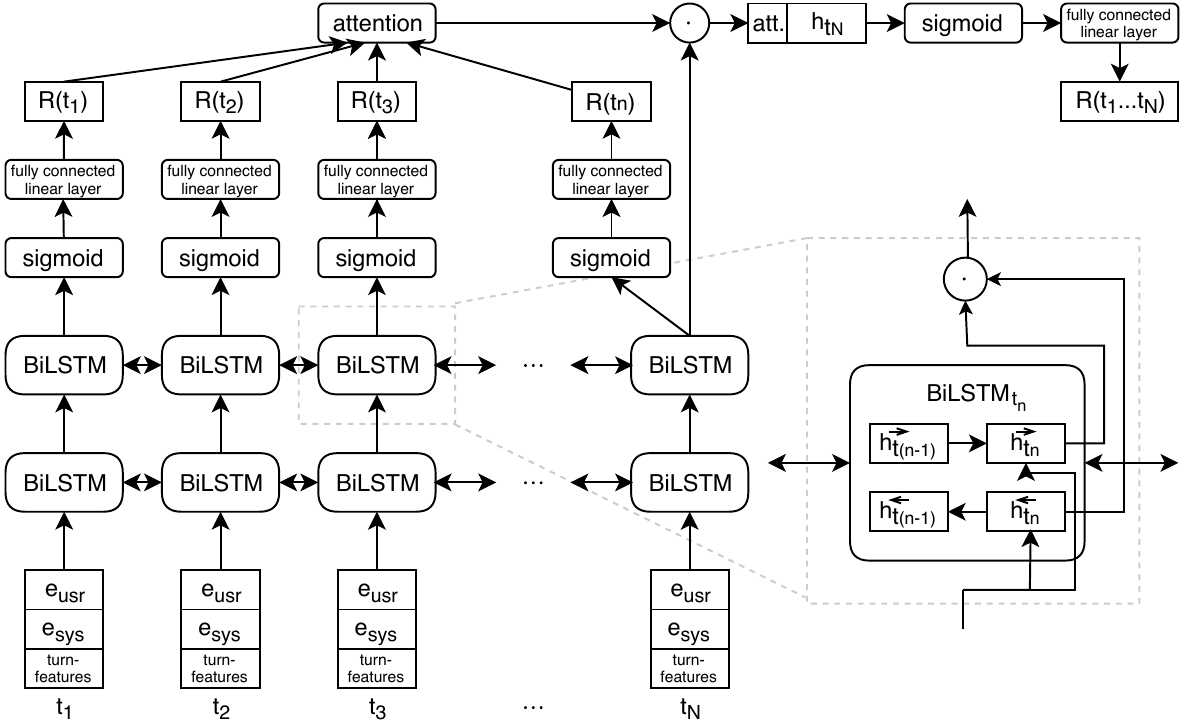}
  \end{flushleft}
  \vspace{-0.2cm}
  \caption{BiLSTM based joint turn and dialogue-level satisfaction estimation model.}
  \label{fig:dialogue-model}
\vspace{-0.5cm}
\end{figure}

\subsection{Multi-task Loss Function for Joint Turn and Dialogue Quality Estimation}
\label{subsubsec:adaptive-loss}
RQ ratings provided by experts are reliable and consistent \citep{Bodigutla_SIGDIAL2019}, however user ratings at the end of a dialogue in general are noisy and it is not clear if they would be cooperative enough to provide correct feedback \citep{DBLP:journals/corr/SuVGKMWY15}. To address the difference in noisiness of labels provided for each task, we followed the approach by \citet{DBLP:journals/corr/KendallGC17} to use homoscedastic (task-dependent) uncertainty to weigh losses from two tasks, where multi-task loss function is derived by maximizing Gaussian likelihood with homoscedastic uncertainty (Equation \ref{eq:1}). Sufficient statistics $\boldsymbol{f^{W}(\pnt{X})}$ is the output of a neural network with weight $\boldsymbol{W}$ on input $\boldsymbol{\pnt{X}}$. $\bm{\pnt{Y_{t}}}$ (turn-ratings) and $\bm{\pnt{Y_{d}}}$ (dialog-ratings) are model outputs.

\vspace{-0.2cm}
\begin{scriptsize}
\begin{equation} \label{eq:1}
\begin{split}
p(\boldsymbol{\pnt{Y_t}},\boldsymbol{\pnt{Y_d}}| \boldsymbol{f^{\pnt{W}}(\pnt{X}})) &= p(\boldsymbol{\pnt{Y_t}}|\boldsymbol{f^{\pnt{W}}(\pnt{X}})) \cdot p(\boldsymbol{\pnt{Y_d}} | \boldsymbol{f^{\pnt{W}}(\pnt{X}})) \\
& =  \mathcal{N}(\boldsymbol{\pnt{Y_t}};  \boldsymbol{f^{\pnt{W}}(\pnt{X})}, \sigma_{t}^2)  \cdot  \mathcal{N}(\boldsymbol{\pnt{Y_d}};  \boldsymbol{f^{\pnt{W}}(\pnt{X})}, \sigma_{d}^2 ) 
\end{split}
\end{equation}
\end{scriptsize}
\vspace*{-\abovedisplayskip}
\vspace*{-\abovedisplayskip}
\begin{scriptsize}
\begin{equation} \label{eq:2}
\vspace{-0.5cm}
\begin{split}
\mathcal{L}(\boldsymbol{\pnt{W}}) &= -\log p(\boldsymbol{\pnt{Y_t}},\boldsymbol{\pnt{Y_d}}| \boldsymbol{f^{\pnt{W}}(\pnt{X}})) \\
& \propto {\dfrac{1}{2\sigma_{t}^2}}||\boldsymbol{\pnt{Y_t}}-\boldsymbol{f^{\pnt{W}}(\pnt{X}})||^2+ ||\boldsymbol{\pnt{Y_d}}-\boldsymbol{f^{\pnt{W}}(\pnt{X}})||^2 + \log \sigma_{t} + \log \sigma_{d}\\
&= {\dfrac{1}{2\sigma_{t}^2}}\mathcal{L}_t(\boldsymbol{W})+ {\dfrac{1}{2\sigma_{d}^2}}\mathcal{L}_d(\boldsymbol{W}) + \log \sigma_{t} + \log \sigma_{d} 	
\end{split}
\vspace{-0.4cm}
\end{equation}
\end{scriptsize}

Equation \ref{eq:2} shows the multi-task loss function $\mathcal{L}$ we minimize. $\mathcal{L}_t$ and $\mathcal{L}_d$ are the mean square error losses computed on turn-level RQ ratings and dialogue-level user ratings respectively. Minimizing the objective functions with respect to noise parameters $\sigma_{t}$ and $\sigma_{d}$ is interpreted as learning the weights for $\mathcal{L}_t$ and $\mathcal{L}_d$ adaptively from the data. Higher the noise, lower is the weight of the corresponding loss. This method to weigh the losses using learnt weights helps in bringing the losses from the two tasks on the same scale as well.  

\section{Data and Experimental Setup}
This section describes our turn and dialogue-level datasets and explains our experimentation setup. 
\subsection{Dialogue Quality Data}
\label{subsec:data}

In order to test the generalizability of the turn and dialogue level user satisfaction models across different domains, user groups and dialogue systems, we sampled 3,129 dialogue sessions (20,167 turns) from $28$ domains (Table \ref{tab:dialogue-stats}). These multi-domain dialogues (Example goals user try to achieve in Appendix Table \ref{tab:user-study-goals}) are representative of end user interactions with Alexa and were randomly sampled from two dialogue systems.

\begin{figure}[!h]
\vspace{-0.1cm}
  \begin{center}
    \includegraphics[width=0.20\textwidth]{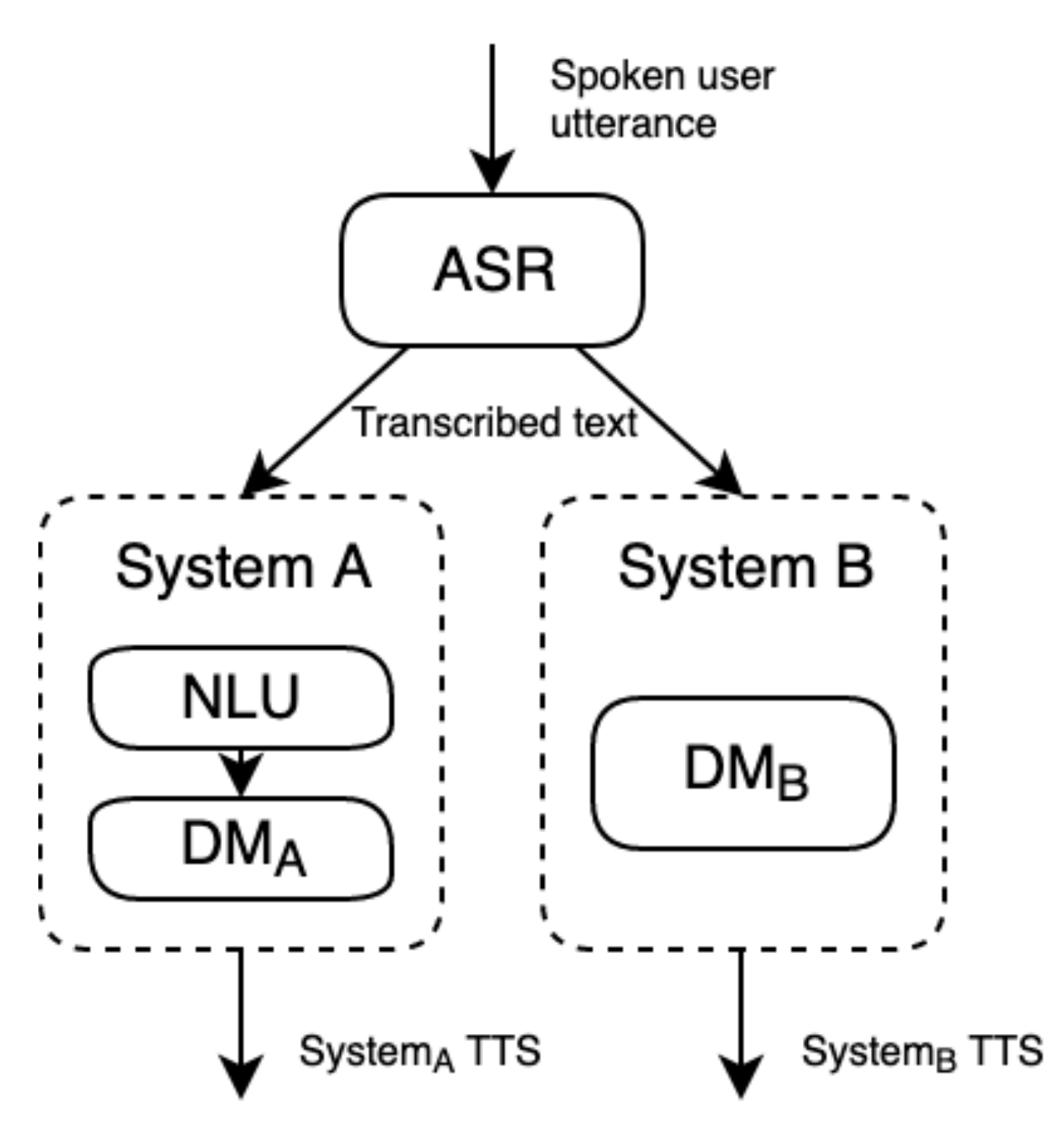}
  \end{center} 
  \vspace{-0.2cm}
  \caption{Dialogue systems A \& B with their own dialogue managers to process user request and generate Text To Speech (TTS) response once the shared ASR component does the speech to text translation.}
  \label{fig:dialogue-systems}
\vspace{-0.3cm}
\end{figure}

Dialogue-system A uses a pipelined modular dialogue agent comprising of ASR, NLU, State Tracker, Dialogue Policy and Natural Language Generation components \citep{Williams2016TheDS}. Dialogue-system B is an end-to-end neural model \citep{ritter-etal-2011-data, shah-etal-2018-bootstrapping} that shares only the ASR component with system A (Fig. \ref{fig:dialogue-systems}).  

\begin{table}
\centering
\resizebox{0.30\textheight}{!}{
 \begin{adjustbox}{max width=\textwidth}
  \bgroup
  \def\arraystretch{1.1}
\begin{tabular}{|c|c|c|c|c|}
\hline
\pbox{10cm} {Dialogue \\ System} &\# Domains & \# Dialogues & \# Turns & \pbox{10cm}{Avg. \# Turns \\per Dialogue}  \\
\hline
A & 24  & 2,133  & 10,774  & 5 \\
\hline
B & 4  & 996  & 9,393  &  9.5 \\
\hline
\end{tabular} 
\egroup
\end{adjustbox}
}
\vspace{-0.1cm}
\caption{Stats on dialogues from dialogue systems A \& B}
\vspace{-0.6cm}
  \label{tab:dialogue-stats}
\end{table}

Each turn was rated by expert RQ annotators\footnote{Expert RQ annotators consistently achieve a high agreement (correlation $>= 0.8$) with other expert annotators and with explicit turn-level user ratings collected through user studies.} and Dialogue level ratings were provided by end users. Users provided their satisfaction rating with the dialogue on a discrete $[1-5]$ scale at the end of each session, irrespective of the outcome. Similar to \citet{Bodigutla_SNEURIPS2019} the rating scale we asked the users to follow was $1$=Very dissatisfied, $2$=Dissatisfied, $3$=Moderately Satisfied (or Slightly dissatisfied), $4$=Satisfied and $5$=Extremely Satisfied. Since earlier attempts to estimate explicit dialogue-level satisfaction ratings did not generalize to different user population (see section \ref{sec:intro}), we collected dialogue ratings from users belonging to ``novice'' ($15\%$), ``some experience'' ($33\%$) and ``experienced'' ($52\%$) groups. A novice user has minimal experience conversing with the SDS and he/she has never used the functionality provided by the $28$ domains prior to the study. A user with some experience has interacted with some (but not all) domains, whereas an experienced user is a seasoned user of Alexa and its domains. 

\subsection{Experimental Setup}
\label{sub:exp-setup}
This section describes the experimental setup we used for training and evaluating turn and dialogue level satisfaction estimation models.

\subsubsection{Turn-level Dialogue Quality Estimation}
\label{subsubsec:exp-setup-turn-level}
Similar to \citet{Bodigutla_SIGDIAL2019}, we considered regression models for experimentation to predict turn-level satisfaction rating on a continuous $[1{-}5]$ scale. We experimented with two variants of the turn-level satisfaction estimation model described in Section \ref{subsec:turn-level-lstm}. In the first variant ($LSTM_{embedding}$) we passed concatenated pre-trained USE sentence embeddings of the user request and system response as input to each time step of the LSTM based model. In the second variant ($LSTM_{embeddings\odot{features}}$) we concatenate USE embeddings with rest of the $15$ turn-level features mentioned in Table \ref{tab:turn-level-features}. We benchmarked the performance of the two LSTM models against the best performing \citep{Bodigutla_SIGDIAL2019} turn-level Gradient Boosting Regression model trained with $48$ hand-crafted features (Appendix Table \ref{tab:features-hand-craft}).

\subsubsection{Dialogue-level Quality Estimation}
\label{subsubsec:exp-setup-dialogue-level}
We experimented with eight models to estimate dialogue level user satisfaction ratings. Three out of the eight models were used as baseline models, which are: 1) Gradient Boosting Regression ($G.Boost$) model trained using features derived from the entire dialogue context ($t_{1:N}$), including hand-crafted turn-level and temporal features (See Appendix Table \ref{tab:features-hand-craft}); 2) Two-layer BiLSTM model ($BiLSTM_{features}$) trained with all turn-level features (Table \ref{tab:turn-level-features}), except for the embeddings themselves; 3) $BiLSTM_{features}$ model with self-attention mechanism ($BiLSTM_{features}^{attn}$), which is also a variant of best performing IQ estimation model  \citep{ultes-2019-improving}. For benchmarking we used best performing \citep{Bodigutla_SNEURIPS2019} $G.Boost_{RQ}$ dialogue-level quality estimation model, which used average predicted RQ rating as an additional feature to train the $G.Boost$ model.

Remaining four models we experimented with comprised of two variants of our proposed BiLSTM based joint dialogue quality estimation model, that used attention over the predicted RQ ratings to predict dialogue level rating (See Section \ref{sec:dialogue-level-lstm}). First variant used only USE embeddings as features ($Joint_{embeddings}^{attn}$) and the second one ($Joint_{embeddings\odot features}^{attn}$) used all the turn-level features mentioned in Table \ref{tab:turn-level-features}.  To test whether including USE embeddings on user request and system response texts alone improved the performance of the baseline $BiLSTM_{features}$ and $BiLSTM_{features}^{attn}$ models, we experimented with their respective counterparts $BiLSTM_{embeddings\odot{features}}$ and $BiLSTM_{embeddings\odot{features}}^{attn}$ models that included USE embeddings as features.

The joint models minimized adaptive weighted loss (Eq. \ref{eq:2}). All the deep neural models we experimented with used Adam \citep{article_adam} optimizer with learning rate $0.0001$, mini-batch size of $64$ and hidden vector size $512$. We used early stopping criteria and ($0.5$) dropout \citep{Srivastava2014DropoutAS} regularization techniques to avoid overfitting. Hyper-parameter ranges we experimented with are in Appendix Table \ref{optimal-hyper-parameter}. 

For both dialogue and turn level quality estimation, dialogues were randomly split into training (80\%), validation (10\%) and test (10\%) sets, so that turns from the same dialogue do not appear in both test and training sets. 

We trained and evaluated the performance of the turn and dialogue-level quality estimation models on dialogues from dialogue-system A and from both systems A \& B combined\footnote{For completeness we evaluated dialogue-quality estimation results using train and test dialogues from System B (Appendix Table \ref{tab:dialogue-quality-results-systemb}). Due to limited data (96 test dialogues) performance comparison between models is inconclusive and needs further experimentation.}. In the first case we used all turn-level features mentioned in Table \ref{tab:turn-level-features}. In the second case we excluded features derived from NLU as dialogue-system B did not use NLU output.

\subsubsection{Evaluation Criteria}

\label{subsubsec:evaluation}
We used Pearson's linear correlation coefficient ($r$) for evaluating each model's 1-5 prediction performance.  For the use case to identify problematic turns from an end user's perspective, it is sufficient to identify {\em satisfactory} (rating $\geq$3) and {\em dissatisfactory} (rating $<3$) interactions \cite{Bodigutla_SNEURIPS2019}.  We used F-score for the dissatisfactory class as the binary classification metric, as most turns and dialogues belong to the satisfactory class. Dialogue-level ratings have a smoother distribution (Pearson's moment coefficient of skewness $-0.27$) over turn-level  RQ ratings (skewness $-0.64$).  

\section{Results and Analysis}
\label{sec:results}
This section presents the turn and dialogue-level user satisfaction estimation results.
\begin{table}[h!]
\centering
\resizebox{0.31\textheight}{!}{
  \begin{adjustbox}{max width=\textwidth}
  \bgroup
  \def\arraystretch{1.3}
\begin{tabular}{|l|c|c|c|c|}
\hline
\multicolumn {1}{|c}{} & \multicolumn{2} {|c|} {System A}  & \multicolumn {2}{|c|} {Systems A \& B }\\
\hline
Model\textbackslash Metric & $Correlation$ & $F-dissat$ & $Correlation$ & $F-dissat.$\\
\hline
Gradient Boosting Regression & 0.77 $\pm$ 0.02 & 0.77 $\pm$ 0.02 & 0.74 $\pm$ 0.02 & 0.72 $\pm$ 0.02  \\
\hline
$LSTM_{embedding}$ & 0.78 $\pm$ 0.02 & 0.80 $\pm$ 0.03 & 0.74 $\pm$ 0.03 & 0.77 $\pm$  0.03 \\
\hline
$LSTM_{embedding\odot{features}}$&  \textbf{0.79 $\pm$ 0.03} & \textbf{0.81 $\pm$ 0.02$\star$} & \textbf{0.76 $\pm$ 0.02} & \textbf{0.78 $\pm$  0.02$\star$} \\
\hline
\end{tabular} 
\egroup
\end{adjustbox}
}
\vspace{-0.2cm}
\caption{Turn-level dialogue quality estimation models' performance measured using correlation between predicted and true ratings and F-score on dissatisfactory class  (F-dissat.) Cells show the mean and 95\% bootstrap confidence interval, highest mean in bold, $\star$ for statistically significant improvement over benchmark Gradient Boosting Regression model's performance.}
\vspace{-0.6cm}
  \label{tab:turn-results}
\end{table}

\subsection{Turn-level satisfaction Estimation}
\label{subsec:turn-results}
As shown in Table \ref{tab:turn-results},  our proposed LSTM based turn-level quality estimation model outperformed the benchmark Gradient Boosting regression model and removed the need to hand-craft features. Even when NLU features were not used, on dialogues from both dialogue systems, the best-performing ($LSTM_{embedding\odot{features}}$) model achieved {\scriptsize $\sim$}3\% relative improvement in correlation ($0.74 \rightarrow 0.76$) and statistically significant (at 95\% boostrap-confidence interval) relative improvement 8.3\% ($0.72 \rightarrow 0.78$) in F-score on dissatisfactory class performance, over the benchmark model.

\subsubsection{Analysis of turn-level model's performance on  new domain}
To further test the generalizability of the $LSTM_{embedding\odot{features}}$ model to new domains, we wanted to verify that the model was not overfitting domain specific vocabulary. To achieve this, we trained the turn-level model with varying percentage of dialogues from a new ``movie reservation \& recommendation'' domain hosted on dialogue System A. Training set consisted of dialogues from Sytems A\&B and specified percentage of dialogues from the new domain\footnote{Since System B did not use NLU, it is not possible to train the model with utterances de-lexicalized using NLU output, such as predicted Intents and slots \cite{tur2011spoken}.}. Consistent with the results in \citep{Bodigutla_SNEURIPS2019}, the prediction performance dropped when no dialogues from the new domain were in the training set (results in Appendix Table \ref{tab:fandango-results}). However, when the model was trained with (randomly sampled) mere 10\% (9\% train, 1\% validation) of dialogues ({\scriptsize $\sim$}6\% slot-value coverage\footnote{Slot-value coverage is the \% of unique (slot-type, value) pairs for the specific domain in the selected set of dialogues.}) , the prediction performance on $F-dissatisfactory$ metric ($0.75 \pm 0.01$) was at par (difference not statistically significant) with the overall performance achieved by the model when it was trained with 90\% (80\% train, 10\% valid) dialogues (Table \ref{tab:turn-results}). Performance parity in-terms of $Correlation$ ($0.74 \pm 0.03$) was achieved when $LSTM_{embedding\odot{features}}$ model was trained with 60\% (54\% train 6\% validation) of dialogues ({\scriptsize $\sim$}50\% slot-value coverage). These two observations imply that binary prediction performance improvement requires training with fewer dialogues in comparison to the number of dialogues required to accurately identify the degree of user (dis)satisfaction. 

In order to further understand the relationship between slot types and annotated RQ labels we calculated the Pointwise Mutual Information (PMI\footnote{PMI of pair of outcomes $(x,y)$ belonging to discrete random variables X,Y is $\log{\frac{p(x,y)}{p(x)p(y)}}$.}) score for the new domain, between its 8 slot-types and 5 RQ labels (total 40 values). Most of the dissatisfactory turns were associated with the system not interpreting the theater names (slot-types `theater') and instructions containing numbers (e.g., ``pick the fourth one'') correctly. Validating our hypothesis that users do not perceive all turns' failures equally, based on the PMI scores, users seem more dissatisfied with system's failure to identify ``theater'' (RQ rating - 1) over failure in interpreting numeric instructions (RQ rating - 2)\footnote{Since RQ ratings are highly correlated with turn-level user satisfaction ratings \citep{Bodigutla_SIGDIAL2019}.}. We calculated cosine similarity between the $40$ dimensional PMI scores vector of ($Slot_{type}$,$RQ_{labels}$) in each selected training set, with PMI scores vector computed on entire set of dialogues in the new domain. As shown in Appendix Table \ref{tab:fandango-results}, the turn-level model's performance on new domain improves with the similarity score. This observation suggests that the model is not overfitting to domain specific vocabulary (e.g., movie name), instead it
 learns the extent of user (dis)satisfaction to failures/success of different (slot) types of requests he/she makes.  

\subsection{Dialogue-level Satisfaction Estimation}
\label{subsec:dialogue-results}

As shown in Table \ref{tab:dialogue-results}, on test sets from System A and System A \& B combined, $Joint_{embeddings\odot features}^{attn}$ model outperformed the seven other models we experimented with. On test dialogues from System A \& B, in comparison to the baseline $BiLSTM_{features}^{attn}$ model, the Joint-model achieved statistically significant (at 95\% confidence interval) absolute 27\% (0.43 $\rightarrow$ 0.70) improvement in correlation and 17\% (0.51 $\rightarrow$ 0.68) in F-score on dissatisfactory class. In comparison to benchmark $G.Boost_{RQ}$ model, the absolute improvement on the same metrics was 7\% and 5\% respectively. Learnt noise ratio of $1.2$ between the two learnt parameters $\sigma_{d}^2$ and $\sigma_{t}^2$ (Eq. $\ref{eq:2}$), shows higher variance in dialogue-level ratings over turn-level labels.

%
\begin{table}[h!]
\vspace{-0.1cm}
\centering
\resizebox{0.31\textheight}{!}{
  \begin{adjustbox}{max width=\textwidth}
  \bgroup
  \def\arraystretch{1.1}
\begin{tabular}{|l|c|c|c|c|c|c|}
\hline
\multicolumn {1}{|c}{} & \multicolumn{2} {|c|} {System A}  & \multicolumn {2}{|c|} {Systems A \& B}\\
\hline
Model\textbackslash Metric &  $Correlation$ & $F-dissatisfactory$  &  $Correlation$ & $F-dissatisfactory$\\
\hline 
$G.Boost$ &  0.59 $\pm$ 0.06 & 0.61 $\pm$ 0.08  & 0.60 $\pm$ 0.05$\dagger$ & 0.63 $\pm$ 0.04$\star\dagger$\\
\hline
$G.Boost_{RQ}$ &  0.66 $\pm$ 0.05$\dagger$ & 0.66 $\pm$ 0.06 &  0.63 $\pm$ 0.06$\star\dagger$ & 0.63 $\pm$ 0.05$\star\dagger$\\
\hline
$BiLSTM_{features}$ &  0.54 $\pm$ 0.07 & 0.63 $\pm$ 0.08  &  0.48 $\pm$ 0.08 & 0.51 $\pm$ 0.06 \\
\hline
$BiLSTM_{embeddings\odot{features}}$ &  0.62 $\pm$ 0.07$\dagger$ & 0.60 $\pm$ 0.07 &  0.66 $\pm$ 0.06$\star\dagger$ & 0.66 $\pm$ 0.05$\star\dagger$\\
\hline
$BiLSTM_{features}^{attn}$&  0.44 $\pm$ 0.10 & 0.51 $\pm$ 0.09 &   0.43 $\pm$ 0.08 & 0.51 $\pm$ 0.06 \\
\hline
$BiLSTM_{embeddings\odot{features}}^{attn}$ &  0.61 $\pm$ 0.08 & 0.64 $\pm$ 0.06 & 0.59 $\pm$ 0.06$\dagger$ & 0.67 $\pm$ 0.04$\star\dagger$ \\
\hline
$Joint_{embeddings}^{attn}$ & 0.68 $\pm$ 0.08$\dagger$ & 0.65 $\pm$ 0.07 & 0.68 $\pm$ 0.06$\star\dagger$ & 0.67 $\pm$0.06$\star\dagger$\\
\hline
$Joint_{embeddings\odot{features}}^{attn}$ &  \textbf{0.69 $\pm$ 0.07$\star\dagger$} & \textbf{0.71 $\pm$ 0.07$\dagger$}  & \textbf{0.70 $\pm$ 0.06}$\star\dagger$ & \textbf{0.68 $\pm$ 0.05}$\star\dagger$\\
\hline
\end{tabular} 
\egroup
\end{adjustbox}
}
\vspace{-0.2cm}
\caption{Performance of dialogue-level quality estimation models\protect\footnotemark. Each cell shows the mean and 95\% bootstrap confidence interval with the highest mean in bold. $\star$ and $\dagger$ indicate statistically significant performance in comparison to baseline $BiLSTM_{features}$ and  $BiLSTM_{features}^{attn}$ models respectively. Compared to Table \ref{tab:turn-results}, wider confidence intervals are due to sparsity of dialogue-level ratings ({\scriptsize $\sim$} 15\% of turn-level ratings).}
\vspace{-0.4cm}
 \label{tab:dialogue-results}
\end{table}
\footnotetext{Results are not broken down further by domain, since a multi-domain conversation session comprises of turns which belong to $\geq$ one domain and context is shared between them.}

Including USE embeddings as features improved the performance of the dialogue-level satisfaction estimation models. Specifically on data from both systems, both $BiLSTM_{embeddings\odot{features}}$ and $BiLSTM_{embeddings\odot{features}}^{attn}$ models achieved around absolute 15\% - 18\% significant improvement in both correlation and F-score on dissatisfactory class performance over their respective counterparts $BiLSTM_{features}$ and $BiLSTM_{features}^{attn}$.

\subsubsection{Analysis of learnt Attention Weights}
For the $Joint_{embeddings\odot features}^{attn}$ model, Table \ref{tab:qual-analysis} shows the attention weights learnt on predicted turn level ($\hat{RQ}$) and true RQ ratings for each turn of a sample dialogue. The joint model puts more weight on the dissatisfactory turns over the satisfactory ones and the dialogue was correctly identified as dissatisfactory. Table \ref{tab:qual-analysis-error} shows an example dialogue where a generous  \citep{Kulikov2018ImportanceOA}  user was satisfied (dialogue-rating $4.0$) even when the system did not offer any alternate time slots or restaurant suggestions when his/her initial request to book a table could not be fulfilled. However, the model predicted the dialogue as dissatisfactory. 


\begin{table}[h!]
  \resizebox{0.31\textheight}{!}{
  \def\arraystretch{2.2}
    \begin{tabular}{ll}
    \toprule
    \multicolumn{1}{c}{\textbf{Utterance}}   & \multicolumn{1}{l}{\textbf{Attention weight and turn-level ratings}} \\
    \midrule
    \textbf{User:} Open \{Anonym. movie app.\}  & \pbox{10cm}{\textbf{System:} Hey this is \{application\} \\\{$W_{attn}:0.073$, $\hat{RQ}:5.0$, ${RQ}:5.0$\}} \\
\textbf{User:} tell me the plot of hustlers  &\pbox{10cm}{\textbf{System:} Here is a summary of Hustlers ... \\ \{$W_{attn}0.077$, $\hat{RQ}:4.3$, ${RQ}:5.0$\}}  \\
\textbf{User:} Give me the whole cast & \pbox{10cm}{\textbf{System:} The main role was played by ... \\ \{$W_{attn}:\textbf{0.087}$, $\hat{RQ}:1.8$, ${RQ}:2.0$\}} \\
\textbf{User:} What is it rated? & \pbox{10cm}{\textbf{System:} It has a rating of 6.6 on i.m.d.b \\ \{ $W_{attn}:0.075$, $\hat{RQ}:4.7$, ${RQ}:5.0$\}}\\
\textbf{User:} Exit & \pbox{10cm}{\textbf{System}: \{Exits the app\} \\ \{$W_{attn}:0.079$, $\hat{RQ}:3.6$, ${RQ}:5.0$\}} \\
\textbf{User:} play hustler soundtrack  & \pbox{10cm}{\medskip \textbf{System:} Heres a station for you \\ based on The Hustler by Willie ...\\\{$W_{attn}:\textbf{0.090}$, $\hat{RQ}:1.1$, ${RQ}:1.0$\}} \\
\textbf{User:} Stop & \pbox{10cm}{\textbf{System:} \{Plays the station\} \\\{$W_{attn}:\textbf{0.090}$, $\hat{RQ}:1.1$, ${RQ}:1.0$\}} \\
\textbf{User:} Open \{Anonym. tkt-booking app\}  & \pbox{10cm}{\textbf{System:} Welcome to \{application\}\\ \{$W_{attn}:0.078$, $\hat{RQ}:4.0$, ${RQ}:5.0$\}} \\
\textbf{User:} Where is hustlers playing ? & \pbox{15cm}{\medskip \textbf{System:} You can go to AMC \\ Which theater would you like? \\ \{$W_{attn}:\textbf{0.087}$, $\hat{RQ}:1.6$, ${RQ}:2.0$\}} \\
\textbf{User:} AMC & \pbox{15cm}{\medskip \textbf{System:} Today at  AMC:  \\Joker...  which movie would you like to see? \\ \{$W_{attn}:\textbf{0.088}$, $\hat{RQ}:1,6$, ${RQ}:2.0$\}}  \\
\textbf{User:} Is the hustlers playing there today & \pbox{10cm}{\medskip \textbf{System:} Today at  AMC: \\ Joker...  which movie would you like to see? \\ \{$W_{attn}:\textbf{0.087}$, $\hat{RQ}:1,7$, ${RQ}:2.0$\}}\\
\textbf{User:} No & \pbox{10cm}{\textbf{System:}\{No response\} \\ \{$W_{attn}:\textbf{0.089}$, $\hat{RQ}:1,3$, ${RQ}:2.0$\}} \\
\bottomrule
\end{tabular}
}
\caption{Conversation\protect\footnotemark  where the $Joint_{embeddings\odot features}^{attn}$ correctly predicts the dialogue as dissatisfactory (rating = $2.5$). $W_{attn}$, $\hat{RQ}$ and true RQ ratings indicate learnt attention weights, predicted and true RQ ratings respectively. Weights in bold indicate top 50\% of learnt attention weights.}
\vspace{-0.2cm}
\label{tab:qual-analysis}
\end{table}
\footnotetext{This dialogue is not a real user conversation with live system.}

\begin{table}[h!]
  \resizebox{0.31\textheight}{!}{
    \def\arraystretch{2.2}
    \begin{tabular}{ll}
    \toprule
    \multicolumn{1}{c}{\textbf{Utterance}}   & \multicolumn{1}{l}{\textbf{Attention weight and turn-level ratings}} \\
    \midrule
    \textbf{User:} Open \{restaurant booking app.\}  & \pbox{10cm}{\textbf{System:} Hey this is \{application\} \\\{$W_{attn}:0.126$, $\hat{RQ}:5.0$, ${RQ}:5.0$\}} \\
\textbf{User:} \pbox{10cm}{Can you can you uh book a \\ table at koreana for two  ?}&\pbox{10cm}{\textbf{System:} Sorry, this is not curr. supported.\\ \{$W_{attn}: \textbf{0.147}$, $\hat{RQ}:1.8$, ${RQ}:2.0$\}}  \\
\textbf{User:} \pbox{10cm}{Can you book a table at the \\ cheesecake factory for three ?} & \pbox{10cm}{\medskip\textbf{System:} What time would you like to \\ make the reservation for? \\ \{$W_{attn}:0.137$, $\hat{RQ}:3.2$, ${RQ}:4.0$\}} \\
\textbf{User:} Three oclock & \pbox{10cm}{\medskip\textbf{System:} What date would you \\ like the reservation for? \\ \{ $W_{attn}:0.135$, $\hat{RQ}:3.6$, ${RQ}:4.0$\}}\\
\textbf{User:} For tomorrow & \pbox{10cm}{\medskip \textbf{System}:  Sorry. No restaurants could \\ be found that match your criteria. \\ \{$W_{attn}: \textbf{0.152}$, $\hat{RQ}:1.1$, ${RQ}:2.0$\}}\\
\textbf{User:} Ok & \pbox{10cm}{\textbf{System:} \{No response\}\\\{$W_{attn}:\textbf{0.153}$, $\hat{RQ}:1.0$, ${RQ}:4.0$\}} \\
\textbf{User:} Stop & \pbox{10cm}{\textbf{System:} \{No response\}\\\{$W_{attn}:\textbf{0.149}$, $\hat{RQ}:1.5$, ${RQ}:4.0$\}} \\
\bottomrule
\end{tabular}
}
\caption{Conversation\protect\footnotemark[9] where the model incorrectly predicts the dialogue as defective (rating $2.3$). User's rating is $4.0$.}
\vspace{-0.7cm}
\label{tab:qual-analysis-error}
\end{table}

\section{Conclusions}
\label{sec:conclusions}
In this paper, we proposed a novel approach to use annotated consistent turn-level Response Quality (RQ) ratings for dialogue level user satisfaction estimation on conversations which span three user groups, $28$ domains and two dialogue systems. With the help of pre-trained Universal Sentence Encoder (USE) embeddings, we removed the need to hand-craft features. Leveraging noise adaptive weighting of multi-task loss technique and aggregating predicted RQ ratings using attention mechanism, we developed the BiLSTM based deep joint turn \& dialogue level satisfaction estimation model. The best-performing joint-model achieved up to 27\% absolute significant improvement in correlation (Pearson's-$r$}) performance and 7\% absolute improvement over the baseline deep neural network and the benchmark $G.Boost$ models, respectively.

Analysis of the learnt attention weights showed that the joint model exhibited the desired behavior to weigh successful and failed turns unequally. However the model was not calibrated to factor in user's preferences and biases, which we plan to address in future work.

\section*{Acknowledgments}
We thank Arindam Mandal and Jean-Jacques Loesch for their guidance and support. We thank Alexa Data Services-RAMP team for data. We also thank EMNLP reviewers for their feedback.

\bibliography{anthology,emnlp2020}
\bibliographystyle{acl_natbib}
\newpage
\appendix
\onecolumn
\section{Appendices}
\label{sec:appendix}
\begin{table*}[h!]
\vspace{-3.5cm}
\captionsetup{font=small}
  \centering
  \resizebox{0.65\textheight}{!}{
  \begin{adjustbox}{max width=\textwidth}
  \bgroup
  \def\arraystretch{1.3}
    \begin{tabularx}{\linewidth}{XX}
    \toprule
    \multicolumn{1}{c}{\textbf{Utterance}}   & \multicolumn{1}{c}{\textbf{Turn Rating}} \\
    \midrule
    \textbf{User:} I would like to book a cab to work  &  \\
    \textbf{System:} Sure, at what time would you like to be picked up?  & \hspace{30mm} \textbf{5.0} \\
    \textbf{User:}  Eight a.m tomorrow &\hspace{4mm}  \\
    \textbf{System:} Could you confirm your cab booking to \{office address\} at 8 a.m tomorrow? 
    			   & \\
		    	   You will be charged \$50 for this ride &\hspace{30mm}  \textbf{5.0} \\
     \textbf{User:}  No, cancel my ride &\hspace{4mm}  \\
    \textbf{System:} Ok, your booking is confirmed.  &\hspace{30mm}  \\
    			   Your card ending xxxx has been charged \$50 & \hspace{30 mm} \textbf{1.0} \\
    \bottomrule
  \end{tabularx}
  \egroup
  \end{adjustbox}
  }
   \caption[Caption for LOF]{An example of a failed dialogue where the overall dialogue rating is \textbf{1.0}.However, average turn ratings \textbf{3.7} indicates a successful dialogue.}
     \vspace{-0.5cm}
  \label{tab:unequal-weighting}
\end{table*}

\begin{table*}[!h]
\vspace{-2.0cm}
\captionsetup{font=small}
\centering
\resizebox{0.65\textheight}{!}{
 \begin{adjustbox}{max width=\textwidth}
  \bgroup
  \def\arraystretch{1.3}
\begin{tabular}{|l|l|}
\hline
 Rating & Description \\
\hline
1 & Terrible (system fails to understand and fulfill user's request)  \\
\hline
2 & Bad (understands the request but fails to satisfy it in any way) \\
\hline
3 &  \pbox{15cm}{OK (understands users request and either partially satisfies the request \\or provides information on how the request can be fulfilled) }\\
\hline
4 &  \pbox{15cm}{Good (understands and satisfies the user request, \\ but provides more information than what the user requested or \\ takes extra turns before meeting the request)}\\
\hline
5 & \pbox{15cm}{Excellent (understands and satisfies user request completely and efficiently}\\
\hline
\end{tabular} 
\egroup
\end{adjustbox}
}
\caption{RQ rating guidelines}
  \label{tab:rq-scale}
\end{table*}
\begin{table*}[h!]
\vspace{-2.5cm}
\captionsetup{font=small}
\scriptsize
  \centering
  \resizebox{0.65\textheight}{!}{
  \begin{adjustbox}{max width=\textwidth}
  \bgroup
  \def\arraystretch{1.3}
  \begin{tabularx}{\linewidth}{ll}
      \toprule
 {\textbf{Goal}} & {\textbf{Domains}} \\
    \midrule
    Get ratings of movies directed by the director of a movie playing in theaters & Movie recommendations and Reservations \\
    Ask for a general type of recipe and then add the ingredients to the shopping list & Recipe, Shopping \\
    Find out the weather in a location and book a ticket to a movie playing in theaters near by & Weather, Location, Movie Recommendations and Reservations \\
    Playing sound track of a popular artist & Knowledge and Music \\
    Book a cab and add a notification for the same & Notifications and Cab booking \\
    Planning activities for eventing & Weather, Restaurants and Cab booking \\    
    \bottomrule   
  \end{tabularx}
  \egroup
  \end{adjustbox}
  }
  \caption{Example goals users tried to achieve and their corresponding domains.}
  \label{tab:user-study-goals}
\end{table*}

\begin{table}[h!]
\captionsetup{font=small}
  \centering
   \resizebox{0.65\textheight}{!}{
  \begin{adjustbox}{max width=\textwidth}
  \bgroup
  \def\arraystretch{1.0}
  \begin{tabularx}{\linewidth}{lll}
    \toprule
  \textbf{Index} & \textbf{Feature set description} & \pbox{3cm}{\textbf{Turn(s) the feature is computed on}} \\
    \midrule
    1& ASR Confidence & \hspace{2mm} $t_{n}^{u}$ \\
    2 &NLU Confidence & \hspace{2mm} $t_{n}^{u}$ \\
    3 &Barge-in & \hspace{2mm} $t_{n}^{u}$ \\
   4& Intent popularity computed on predicted NLU intent &  \hspace{2mm}$t_{n}^{u}$ \\
  5 & Domain popularity computed on predicted NLU intent & \hspace{2mm}$t_{n}^{u}$ \\
   6 & NLU Intent similarity between consecutive turns & \hspace{2mm} $t_{n}^{u}$-$t_{n+1}^{u}$ \\ 
  7 & Syntactic similarity between consecutive turns user utterances & \hspace{2mm} $t_{n}^{u}$-$t_{n+1}^{u}$ \\ 
 8  & Syntactic similarity between user utterance \& system response & \hspace{2mm} $t_{n}^{u}$-$t_{n}^{s}$ \\ 
   9& \pbox{10cm}{Syntactic similarity between current response \\\& previous turn's system response} & \hspace{2mm} $t_{n-1}^{s}$-$t_{n}^{s}$ \\ 
  10 & Affirmation prompt in user request & \hspace{2mm} $t_{n}^{u}$ \\
  11 & Negation prompt in user request & \hspace{2mm} $t_{n}^{u}$ \\
  12 & Question prompt in user request & \hspace{2mm} $t_{n}^{u}$ \\
  13 & Termination prompt in user request & \hspace{2mm} $t_{n}^{u}$ \\
   14& Next turn's ASR Confidence & \hspace{2mm} $t_{n+1}^{u}$ \\
   15& Next turn's NLU Confidence & \hspace{2mm} $t_{n+1}^{u}$ \\
   16 &Next turn's Barge-in indicator & \hspace{2mm} $t_{n+1}^{u}$ \\
  17 & Affirmation prompt in next turn's user request & \hspace{2mm} $t_{n+1}^{u}$ \\
   18 & Negation prompt in next turn's user request & \hspace{2mm} $t_{n+1}^{u}$ \\
  19 & Question prompt in next turn's user request & \hspace{2mm} $t_{n+1}^{u}$ \\
   20 & Termination prompt in next turn's user request & \hspace{2mm} $t_{n+1}^{u}$ \\
  21 & Intent popularity computed on next turn's predicted NLU intent &  \hspace{2mm}$t_{n+1}^{u}$ \\
  22 & Domain popularity computed on next turn's predicted NLU intent & \hspace{2mm}$t_{n+1}^{u}$ \\
  23 & Affirmation prompt in system response & \hspace{2mm} $t_{n}^{s}$ \\
  24 & Negation prompt in system response & \hspace{2mm} $t_{n}^{s}$ \\
  25 & Question prompt in system response & \hspace{2mm} $t_{n}^{s}$ \\
  26 & Un-actionable user request & \hspace{2mm} $t_{n}^{s}$ \\
  27 & \# Un-actionable user request & \hspace{2mm} $t_{1}^{s}$-$t_{n}^{s}$ \\
  28 & \# Barge-ins & \hspace{2mm} $t_{1}^{u}$-$t_{n}^{u}$ \\
 29 &   \# Question prompt in system response & \hspace{2mm} $t_{n}^{s}$ \\
 30 &  \# Negation prompt in system response & \hspace{2mm} $t_{n}^{s}$ \\
  31 & \# Affirmation prompt in system response & \hspace{2mm} $t_{n}^{s}$ \\
  32 & \# Termination prompt in user request & \hspace{2mm} $t_{n}^{u}$ \\
  33 & \#  Question prompt in user request & \hspace{2mm} $t_{n}^{u}$ \\
  34 &  \# Negation prompt in user request & \hspace{2mm} $t_{n}^{s}$ \\
  35 & \# Unique Intents/\# Length of dialogue so far & \hspace{2mm} $t_{1}$-$t_{n}$ \\
   36  &   Length of the dialogue so far & \hspace{2mm} $t_{1}$-$t_{n}$ \\
   37 & Avg ASR confidence &\hspace{2mm} $t_{1}^{u}$-$t_{n}^{u}$ \\ 
  38 & Avg NLU confidence &\hspace{2mm} $t_{1}^{u}$-$t_{n}^{u}$ \\ 
  39 & Avg Semantic similarity between consecutive turns' user utterances &\hspace{2mm} $t_{1}^{u}$-$t_{n+1}^{u}$ \\ 
  40 & Avg Syntactic similarity between consecutive turns' user utterances &\hspace{2mm} $t_{1}^{u}$-$t_{n+1}^{u}$ \\ 
  41 & Avg Semantic similarity between consecutive turns' system responses &\hspace{2mm} $t_{1}^{s}$-$t_{n}^{s}$ \\ 
  42 & Avg Syntactic similarity between consecutive turns' system responses &\hspace{2mm} $t_{1}^{s}$-$t_{n}^{s}$ \\ 
  43 & Avg Semantic similarity between user utterance and system responses &\hspace{2mm} $t_{1}^{u}$-$t_{n}^{s}$ \\ 
  44 & Avg Syntactic similarity between user utterance and system responses &\hspace{2mm} $t_{1}^{u}$-$t_{n}^{s}$ \\ 
   45 & Avg aggregate - domain popularity & \hspace{2mm} $t_{0}$-$t_{n}$ \\
   46 & Avg time difference between consecutive utterances & \hspace{2mm} $t_{0}^u$-$t_{n}^u$  \\
   47 &Avg aggregate - intent popularity & \hspace{2mm} $t_{0}$-$t_{n}$ \\
   48 & Avg aggregate - domain popularity & \hspace{2mm} $t_{0}$-$t_{n}$ \\
   \bottomrule
  \end{tabularx}
  \egroup
  \end{adjustbox}
  }
   \caption{Features used for predicting turn ratings by \citet{Bodigutla_SIGDIAL2019}. \# indicates count. Features 10-45 cover dialogue system specific rule based turn-level features and hand-crafted temporal features.}
  \label{tab:features-hand-craft}
\end{table}

\pagebreak

\begin{table*}[h!]
\captionsetup{font=small}
\centering
    \resizebox{0.65\textheight}{!}{
      \bgroup
  \begin{adjustbox}{max width=\textwidth}
  \def\arraystretch{1.2}
  \begin{tabularx}{\textwidth}{X X}
    \toprule
    \multicolumn{1}{c}{\textbf{Model}}  & \multicolumn{1}{c}{\textbf{Hyper parameter and their corresponding ranges}}  \\
    \midrule
   Gradient Boosting Decision Trees & \pbox{18cm}{max-depth:\,$[2{-}10]$,\\ \,min-samples-leaf:\,$[2{-}10]$, \\ \,min-samples-split:\,$[2{-}10]$ } \\\\
  \pbox{10cm} {LSTM and BiLSTM Based models  \\ for turn and dialogue level quality estimation} & \pbox{18cm}{n-layers:\,$[1,2,3]$,\\ \,hidden size:\,$[8,16,32,64,128,256,512,1024]$,  \\ \, batch size:\,$[8,16,32,64,128]$, \\ \,optimization:\,$[sgd, Adam, RMSProp]$, \\ \, dropout: $[0.1, 0.2, 0.3, 0.4, 0.5, 0.6, 0.7]$, \\ \,learning rate:\,$[1.0, 0.1, 0.001, 0.0001, 0.00001]$, \\ \,length of the sequence:\,$[9-20]$ }  \\\\
  \bottomrule
  \end{tabularx}
  \end{adjustbox}
  \egroup
  }
  \vspace{-0.1cm}
   \caption{Hyper parameter value ranges we used for training turn-level and dialogue-level quality estimation. DNN models were implemented in PyTorch \citep{NEURIPS2019_9015} and Gradient Boosting Regression model was implemented using scikit-learn \citep{CarolynC2000}. Parameters were tuned using grid-search and experiments were run on  P2.xlarge AWS EC2 compute instance, which has 1 NVIDIA K80 GPU, 4 vCPUs, 61GiB RAM.}
  \label{optimal-hyper-parameter}
\end{table*}

\begin{table*}[h!]
\captionsetup{font=small}
\vspace{0.1cm}
\centering
\resizebox{0.65\textheight}{!}{
  \begin{adjustbox}{max width=\textwidth}
  \bgroup
  \def\arraystretch{1.1}
\begin{tabular}{|l|c|c|c|}
\hline
\multicolumn {1}{|c}{} & \multicolumn{2} {|c|} {System B} \\
\hline
Model\textbackslash Metric & $Correlation$ & $F-dissatisfactory$ \\
\hline 
$G.Boost$ & 0.67 $\pm$ 0.10 $\dagger$ & 0.64 $\pm$ 0.07\\
\hline
$G.Boost_{RQ}$& 0.70 $\pm$ 0.09  $\dagger$  & 0.64 $\pm$ 0.06 \\
\hline
$BiLSTM_{features }$ &  0.53 $\pm$ 0.13 & 0.66 $\pm$ 0.11 \\
\hline
$BiLSTM_{embeddings\odot{features}}$   & 0.69 $\pm$ 0.09  $\dagger$  & 0.66 $\pm$ 0.10 \\
\hline
$BiLSTM_{features}^{attn}$ &   0.38 $\pm$ 0.15  & 0.57 $\pm$ 0.14  \\
\hline
$BiLSTM_{embeddings\odot{features}}^{attn}$ & 0.68 $\pm$ 0.11 $\dagger$  & 0.71 $\pm$ 0.08  \\
\hline
$Joint_{embeddings}^{attn}$ &  0.62 $\pm$ 0.10 & 0.72 $\pm$  0.08 \\
\hline
$Joint_{embeddings\odot{features}}^{attn}$ & 0.65 $\pm$ 0.10  $\dagger$  & 0.65 $\pm$  0.09 \\
\hline
\end{tabular} 
\egroup
\end{adjustbox}
}
\vspace{-0.1cm}
\caption{Dialogue-level quality estimation on 96 test dialogues from System B. Dialogues from training (800) and validation (100) were obtained from the same system as well. Models were trained without NLU features. Larger 95\% bootstrap confidence intervals around the mean are due to limited test data. $\dagger$  indicates statistical significance over $BiLSTM_{features}^{attn}$  model's results.}
  \label{tab:dialogue-quality-results-systemb}
\end{table*}

\begin{table*}[h!]

\captionsetup{font=small}
\centering

\resizebox{0.65\textheight}{!}{
  \begin{adjustbox}{max width=\textwidth}
  \bgroup
  \def\arraystretch{1.1}
\begin{tabular}{|l|c|c|c|c|}
\hline

\%Train Dialogues& $F-dissatisfactory$ & $Correlation$ & Slot-Value coverage & $Cos\_sim(PMI^{train}_{(slot-type, label)}, PMI^{all-dialogues}_{(slot-type, label)})$\\
\hline 
$0$ &  0.68 $\pm$ 0.02  & 0.55 $\pm$ 0.03 & - & - \\
\hline
$10$ &  0.75 $\pm$ 0.01 & 0.67 $\pm$ 0.02 & 6.20\% & 0.516\\
\hline
$20$ & 0.74 $\pm$ 0.02  & 0.68 $\pm$ 0.02 & 13.5\% & 0.546\\
\hline
$30$ & 0.76 $\pm$ 0.02 & 0.69 $\pm$ 0.02 & 21.1\% & 0.781\\
\hline
$40$ &  0.78 $\pm$ 0.02 & 0.73 $\pm$ 0.02 & 31.4\% & 0.796\\
\hline
$50$ &   0.77 $\pm$ 0.03  & 0.72 $\pm$ 0.03  & 40.9\%& 0.854\\
\hline
$60$   & 0.79 $\pm$ 0.03  & 0.74 $\pm$ 0.03  & 48.9\% & 0.864\\
\hline
$70$ &  0.80 $\pm$ 0.02 & 0.74 $\pm$  0.03  & 58.8\% & 0.886\\
\hline
$80$ &  0.83 $\pm$ 0.04  & 0.77 $\pm$  0.03 & 70.9\% & 0.931\\
\hline
$90$ & 0.84 $\pm$ 0.04 & 0.78 $\pm$  0.03 & 83.9\% & 0.963\\
\hline
\end{tabular} 
\egroup
\end{adjustbox}
}
\vspace{-0.1cm}
\caption{Turn-level $LSTM_{embeddings\odot{features}}$ model's performance on multi-domain dialogues consisting of new multi-turn Movie Reservation \& Recommendation domain (450 dialogues, 1500 turns). Train dialogues \% indicates, the percentage of dialogues (out of 450) used for training (90\% train , 10\% validation split). Slot-Value coverage is the percentage of unique (slot-type, value) pairs in each training set. $Cos\_sim$ is the cosine similarity between 40 (8 slot-type x 5 RQ-label categories) dimensional Pointwise Mutual Information (PMI) vectors computed on (slot-type, label) pair from dialogues in training set with PMI vector computed on (slot, label) pair from all 450 dialogues.}

  \label{tab:fandango-results}
\end{table*}
\clearpage

\end{document}